\documentclass{article}
\usepackage{spconf,amsmath,graphicx}
\usepackage{balance}
\usepackage{url}

\usepackage{amsmath,amssymb,multirow}
\usepackage{caption}
\title{diffDeMorph: Extending Reference-Free Demorphing to Unseen Faces}

\name{Nitish Shukla and Arun Ross}
\address{Michigan State University, USA}
\begin{document}
%
\maketitle
\begin{abstract}
A face morph is created by combining two face images corresponding to two identities  to produce a composite that successfully matches both the constituent identities. Reference-free (RF) demorphing reverses this process using only the morph image, without the need for additional reference images. Previous RF demorphing methods are overly constrained, as they rely on assumptions about the distributions of training and testing morphs such as the morphing technique used (e.g., landmark-based) and face image style (e.g., passport photos). In this paper, we introduce a novel diffusion-based approach, referred to as diffDeMorph,  that effectively disentangles component images from a composite  morph image with high visual fidelity. Our method is the first to generalize across morph techniques and face styles, beating the current state of the art by $\geq 59.46\%$ under a common training protocol across all datasets tested. We train our method on morphs created using synthetically generated face images and test on real morphs, thereby enhancing the practicality of the technique. Experiments on six datasets and two face matchers establish the utility and efficacy of our method.   
\end{abstract}
\begin{keywords}
morphs, face demorphing, biometrics, DDPM
\end{keywords}
\section{Introduction}
\vspace{-0.3cm}

A face morph is created by combining two face images corresponding to two different identities to produce a composite image that exhibits high biometric similarity to each of the constituent identities \cite{ref27}.
Such images can be used to launch morph attacks, where, for example, a single ID card may be shared by multiple individuals. Traditionally, face morphs have been created by first extracting landmarks from faces and then aligning them for composition  \cite{ref23,ref21}. However, more recently, deep-learning based techniques have been successfully used  in generating high-quality morphs \cite{ref9,ref13,ref14}. 
\begin{figure}[ht!]
    \centering
    \includegraphics[width=0.8\linewidth]{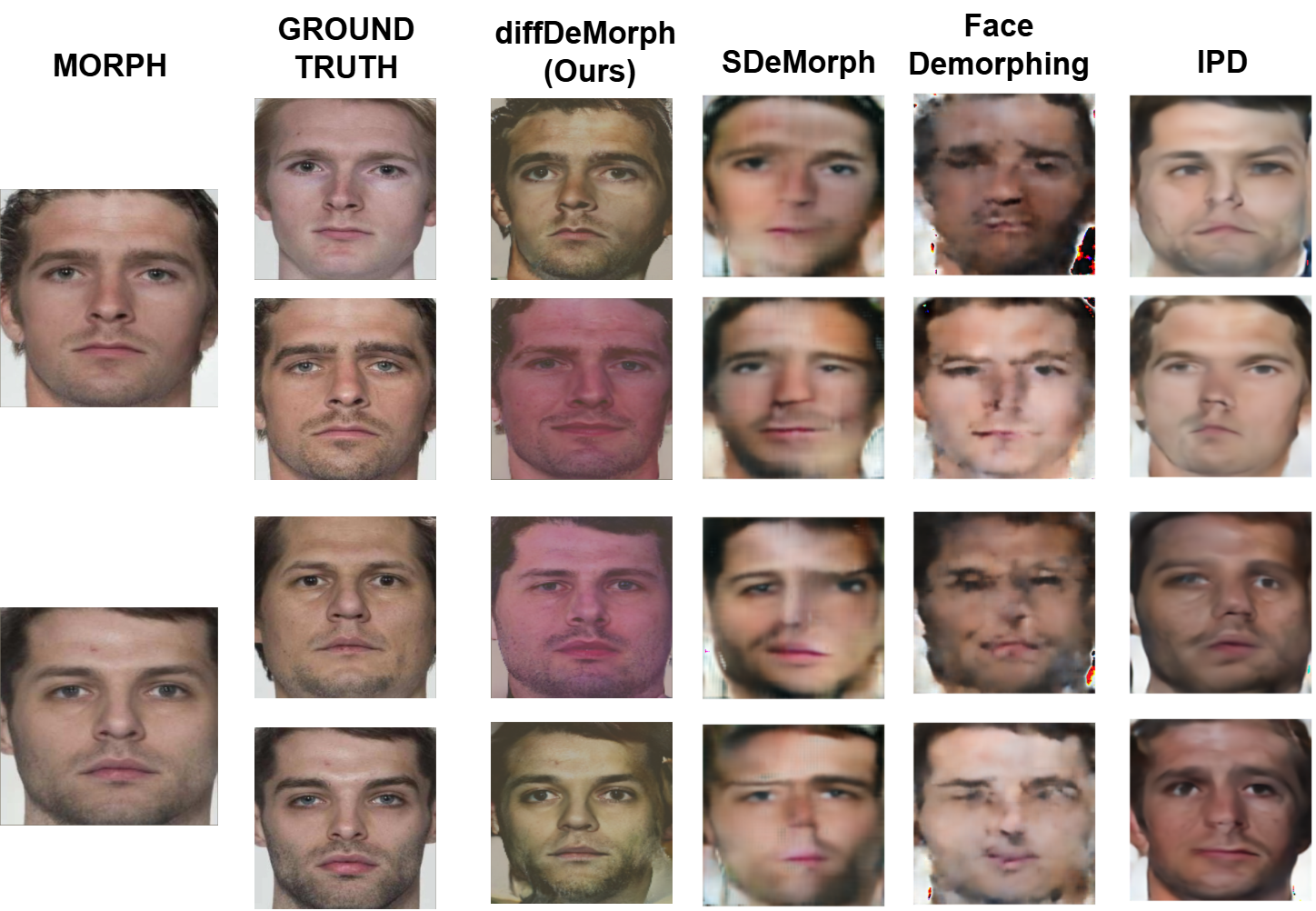}
    \caption{Comparing diffDeMorph with existing reference-free demorphing methods under a common training protocol. Our method produces significantly better results in terms of visual fidelity and biometric match to the ground truth images. Note that our training protocol reflects a viable real-world use case. }
    \label{fig:comparison}
\end{figure}

Facial demorphing is the inverse of this process. The goal is to recover the images used to create the morphs. Although Morph Attach Detection (MAD) techniques can flag  morph images, they do not reveal any information about the images used to create the morph. Hence, the need for demorphing arises. Original facial demorphing methods were reference-based (RB) \cite{ref16,ref53}, i.e., they require an image of one of the constituent identities in addition to the morph to recover the second image. More recently, reference-free (RF) demorphing methods, which require only the morph image, have been proposed with varying degrees of success \cite{ref18,ref51,ref66}. RF demorphing is an inherently complex task due to the lack of constraints in image space as well as the absence of prior information such as the morph technique used. Indeed, given a single image, there are an infinite number of possible decompositions. Thus, the goal is to learn an applicable prior that can be generalized across morph distributions. 
\label{sec:intro}

Due to inherent complexities, previous methods for RF demorphing \cite{ref18,ref51,ref66} operated under strict assumptions. While \cite{ref18} and \cite{ref66} assumed that train and test morphs are created using a common set of face images, \cite{ref51} assumed that train and test morphs have similar passport-style face images. Moreover, all three methods assumed that the test morphs are generated using the same morph technique as the training morphs. Such assumptions significantly limit the generalization of these methods, leading to poor performance in real-world demorphing scenarios.
In \cite{ref18}, the authors proposed a diffusion-based method for demorphing. Although their method uses diffusion during the training phase, the backward sampling during inference lacks any generative aspect, which essentially reduces the denoising network to an equivalent auto-encoder. Moreover, they also assume that the train and test morphs are created from a common pool of face images, which is a strong assumption limiting the applicability of their method.

In this work, we address the aforementioned limitations by  i) introducing a morph-guided denoiser which conditions the denoising process on the morph image in the RGB domain, and ii) relaxing any constraints on the distribution of train/test morphs. Our method operates under a considerably more challenging and realistic protocol.

In summary, our contribution is as follows:
\begin{itemize}
    \item We propose a novel diffusion-based demorphing method, referred to as diffDeMorph,  which reconstructs the constituent images conditioned on the morph in the RGB domain. Our method outperforms the current state of the art by a significant margin under a common training protocol.
    \item To the best of our knowledge, our method is the first to perform demorphing with no assumptions about image style/morphing technique used to create test morphs, thereby enhancing the practical utility of the method. 
    \item We further explore various scenarios arising in the real world in terms of train/test distribution (see Section \ref{sec:results}). Our method consistently outperforms existing methods with excellent visual fidelity.
\end{itemize}

\vspace{-0.3cm}
\section{BACKGROUND }
\vspace{-0.3cm}
\label{sec:format}
\subsection{Face Demorphing}
\vspace{-0.2cm}
\label{back:fd}
Throughout this paper, we denote the morph image as $x$ and the constituent images as $i_1,i_2$ such that
$x=\mathcal{M}(i_1,i_2)$,
where, $\mathcal{M}$ is a black-box morphing operator. The goal of $\mathcal{M}$ is to ensure that $\mathcal{B}(x,i_k)>\tau$, $k\in \{1,2\}$, with respect to a face matcher $\mathcal{B}$ that produces similarity scores  and a threshold $\tau$. Demorphing attempts to invert the morphing process. A demorphing operator, $\mathcal{DM} (=\mathcal{M}^{-1})$, inputs the morph image, $x$, and reconstructs the  constituent images, $o_1,o_2=\mathcal{DM}(x)$,
satisfying the conditions:
\begin{equation}
\label{eq3}
\mathcal{B}(o_1,o_2)<\theta
\end{equation}
\begin{equation}
\label{eq4}
    \min_{j\in\{1,2\}} \max_{\substack{k \in \{1,2\} \\ k \neq j}} \{\ \ \mathcal{B}(o_j,i_k),\mathcal{B}(o_j,i_j)\ \ \} >\epsilon
\end{equation}
where, $\theta$ and  $\epsilon$ are similarity thresholds.
Eqn. (\ref{eq3}) enforces the reconstructed outputs to appear dissimilar to each other (to avoid the morph replication problem \cite{dcgan}) and Eqn. (\ref{eq4}) ensures that each reconstructed output matches with its corresponding ground truth image.

In the literature, demorphing has been explored under various scenarios based on the composition of the training and testing morphs \cite{ref66,dcgan}. 
Consider a set of face images, $\mathcal{Y}$, that is used to generate the train and test morphs.
Scenario 1: Train and test morphs are generated from pairs of identities in $\mathcal{Y}$. This means that some train and test morphs will be created from the same identity pairs.
Scenario 2: Here, $\mathcal{Y}$ is divided into two sets with disjoint identities, $\mathcal{Y}_1$ and $\mathcal{Y}_2$. The train morphs are created from pairs of identities in $\mathcal{Y}_1$. Each test morph is created using one identity in $\mathcal{Y}_1$ and another identity in $\mathcal{Y}_2$.
Scenario 3: Here, $\mathcal{Y}$ is divided into two sets with disjoint identities, $\mathcal{Y}_1$ and $\mathcal{Y}_2$. The train morphs are created from pairs of identities in $\mathcal{Y}_1$. The test morphs are created from pairs of identities in $\mathcal{Y}_2$.
Note that the datasets used in this paper consist of a single face image per identity, with only neutral face images from the FRLL dataset being used to generate the test morphs. Consequently, the terms ``images" and ``identities" are sometimes used interchangeably.



\label{eq:scenario}
In this paper, we focus on scenario 3 making the problem considerably more challenging compared to previous works.
\begin{figure}
    \centering
    \includegraphics[width=0.7\linewidth]{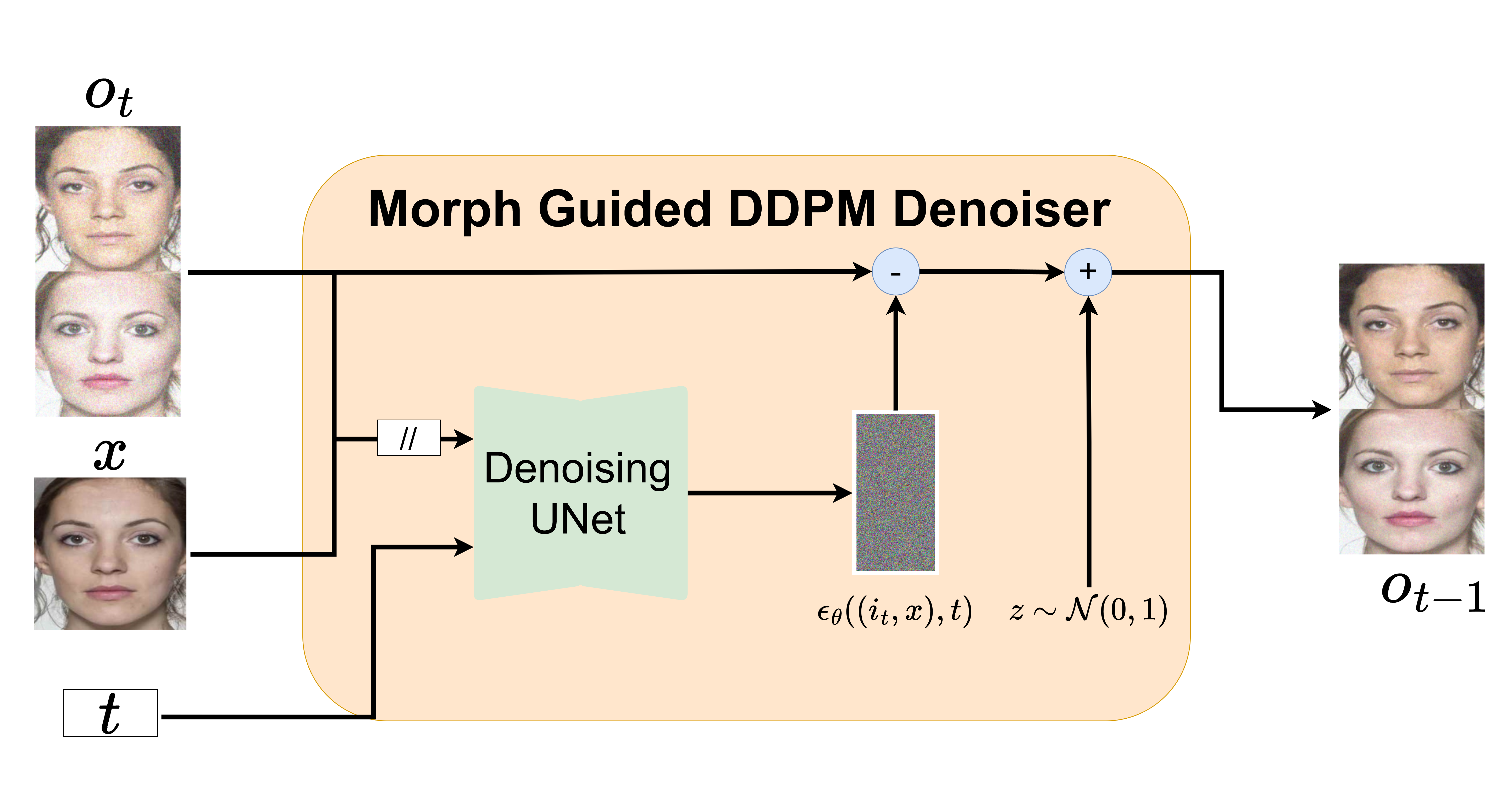}
    \caption{Architecture of our morph guided denoiser. During inference, the morph is appended to the noisy output sample $o_t$, providing the guidance for generation of $o_{t-1}$. This process is repeated until $o=(o_1,o_2)_0$ are recovered.   }
    \label{fig:arch}
\end{figure}
\vspace{-0.2cm}
\subsection{DDPM}
\vspace{-0.2cm}
On a high level, Diffusion Denoising Probabilistic Models (DDPMs)  \cite{ref108} are generative models characterized by a forward noise-inducing process and a backward denoising process. The forward process is a Markov chain based on a fixed variance schedule $\{\beta_t\in(0,1)\}_{t=0}^T$. Given a training dataset $x_0\sim q(x_0)$, the forward diffusion process adds Gaussian noise to the data in $T$ timesteps:
\vspace{-0.5cm}
\begin{center}
   \begin{equation}
    q(x_t|x_{t-1})=\mathcal{N}(x_t;\sqrt{1-\beta_t}x_{t-1},\beta_t\mathbb{I})
\end{equation} 
\end{center}

The diffusion probability, $q(x_t|x_0)$, and the sampling probability, $q(x_{t-1}|x_0)$,  can be expressed using Bayes’ rule and Markov property as: 
    \begin{equation}
        q(x_t|x_0)=\mathcal{N}(x_t;\sqrt{\Bar{\alpha}}x_0;(1-\Bar{\alpha})\mathbb{I}) ,\ \ t=1,2,...,T
    \end{equation}
    \begin{equation}
        q(x_{t-1}|x_t)=\mathcal{N}(x_{t-1};\Bar{\mu}(x_t,x_0);\tilde{\beta}\mathbb{I}) ,t=1,2,...,T
    \end{equation}

where, $\alpha_t=1-\beta_t$, $\Bar{\alpha}_t=\prod_{s=1}^t\alpha_s$, $\tilde{\beta}=\frac{1-\Bar{\alpha}_{t-1}}{1-\Bar{\alpha}_t}\beta_t$ and 

$\tilde{\mu}(x_t,x_0)=\frac{\sqrt{\Bar{\alpha}_t}\beta_t}{1-\Bar{\alpha}_t}x_0 + \frac{\sqrt{\alpha_t}(1-\Bar{\alpha}_{t-1})}{1-\Bar{\alpha}_t}x_t $.

DDPM constructs a Markov chain through a reverse diffusion process, beginning with a prior distribution defined as \( p(x_T) = \mathcal{N}(x_T; 0, \mathbb{I}) \). The reverse process employs a Gaussian transition distribution, expressed as:
\vspace{-0.3cm}
\[
p(x_{t-1} \mid x_t) = \mathcal{N}(x_{t-1}; \mu_\theta(x_t, t), \Sigma_\theta(x_t, t)),
\]
where \( \mu_\theta(x_t, t) \) and \( \Sigma_\theta(x_t, t) \) denote the mean and variance predicted by the model, parameterized by \( \theta \).
The parameters \( \theta \) are optimized to ensure that the generated reverse process closely approximates the noise introduced during the forward process. The training objective is designed to minimize a loss function, which has a closed form as the KL divergence between Gaussian distributions. This objective can be simplified to $ \mathbb{E}_{x, \epsilon \sim \mathcal{N}(0,1), t} \left[ \| \epsilon - \epsilon_\theta(x_t, t) \|_2^2 \right]$,
where, \( \epsilon_\theta(x_t, t) \) is the noise predicted by the model at time step $t$ and $\epsilon$ is the actual noise added in the forward process.
\vspace{-0.3cm}
\section{PREVIOUS WORK}
\vspace{-0.3cm}
\label{sec:pagestyle}
In one of the earliest works \cite{ref51}, the authors used GANs to decompose the morph into constituent images. Their method employed an image-to-image generator and three Markovian discriminators. While their method worked under scenario 3 (see section \ref{back:fd}), they still assumed that train and test morphs have similar passport-style faces. Moreover, their method also tends to produce outputs closely resembling the morph itself. Shukla \cite{ref18} proposed a diffusion-based method for scenario 1 that iteratively adds noise to the morph image and recovers the constituent face images during the denoising process. This paper extends this work to include scenario 3. In \cite{ref66}, authors proposed a method for  scenario 1 that decomposes the input morph into multiple unintelligible components using a decomposer network. A merger network weighs and combines the components to recover the constituent images.
\vspace{-0.3cm}
\section{Methodology}
\vspace{-0.3cm}



We pose demorphing as an iterative, coupled image denoising problem. In RGB domain, the constituent images $i_1$ and $i_2$ exist independently, and the morph image, $x$, establishes the relationship between them through  the triplet $(x,i_1,i_2)$. Consequently, it is reasonable to conceptualize these images as a single paired entity, $i=(i_1,i_2)$, where generation/demorphing is conditioned on the morph. Traditionally, demorphing is treated as a one-to-many one-shot image generation task where each generated output is individually conditioned on the morph, often resulting in the outputs looking similar to each other. However, we formulate the generation as a one-to-one function from the RGB morph domain to the coupled face image space. The main idea is to generate the outputs as one object from noise, where the morph image is injected as guidance in each generation step. We employ DDPMs \cite{ref108} to capture the joint distribution of the constituent images, conditioned on the morph.

In \cite{ref18}, the authors used DDPM to disentangle the constituent images from the morph. However, their method has two main limitations; they assume that train and test morphs are i) made using the same morphing technique and ii) share the same pool of constituent images (scenario 1). In this paper, we extend \cite{ref18} to the much more challenging scenario 3. Moreover, we also relax the assumption of having the same morph technique being used to produce train and test morphs. Note that this reflects a viable real-world use case. While \cite{ref18} uses diffusion in the training phase, the evaluation routine lacks any kind of backward sampling process due to the mismatch in input-output dimensions, which essentially reduces their denoising network to an equivalent autoencoder. This severely limits their method to demorph unseen morph distributions. We remedy these limitations by introducing morph-guided coupled-diffusion.
\vspace{-0.2cm}
\subsection{Coupled Forward Diffusion}
\vspace{-0.2cm}
The coupled forward diffusion incrementally adds a small amount of Gaussian noise to the coupled ground truth input $i=(i_1,i_2)$, in steps ranging from 0 to $T$  resulting in the noisy sequence  $(i_0, i_1,i_2,..., i_T)$, where, $i_0$ is the unmixed sample and $i_T$ tends to an isotropic Gaussian distribution as $T\rightarrow\infty$. The noise added during each step is usually fixed and controlled by a fixed variance schedule $\{\beta_t\in(0,1)\}_{t=0}^T$. During the coupled forward process, we add noise to the coupled input $(i_1,i_2)$ until it degenerates
 into pure Gaussian noise.
\vspace{-0.2cm}
\subsection{Morph Guided Reverse Sampling}
\vspace{-0.2cm}
The goal of the learning is to estimate the coupled face images conditioned on the morph image, i.e., $p((i_1,i_2)|x)$.  We follow a similar setup as in \cite{ref108}. We integrate the control condition (morph image) by performing supervised training with the loss function: 
$\mathcal{L}=\mathbb{E}_{x,i\sim p(i|x),\epsilon,t} \left[ || \epsilon-\epsilon_\theta((i_t,x),t)||_2^2\right]$
where, $x$ is the morph image and $i_t=(i_1,i_2)_t$ is the noisy coupled sample at time step $t$. It is important to note that, unlike conventional conditioning approaches, we provide the conditioning information in the RGB domain alongside the input to the denoising network. This strategy also enables the denoising process to focus on less perceptually prominent features, such as hair, background and other details, which may be overlooked when using compressed conditioning features, as typically employed in standard text-to-image diffusion methods. 
During inference, the denoising network concatenates the morph image $x\in\mathbb{R}^{3\times h\times w}$, with the noise sample $o_t\in\mathbb{R}^{6\times h\times w}$ at time step $t$ to recover  $o_{t-1}\in\mathbb{R}^{6\times h\times w}$. This process is repeated until the outputs at $t=0$ are recovered. We visualize this process in Figure \ref{fig:arch}.

\textbf{Implementation Details:} In all our experiments, we set $T = 1000$ for training. During sampling, we set the number of steps to 100. The beta schedule for variances in the forward process is scaled linearly from $\beta_0=10^{-4}$ to $\beta=0.02$.  The training was done
 using Adam optimization with an initial learning rate of $10^{-3}$ for 300 epochs. For the denoising network, we use UNet \cite{ref19} with 9 input channels and 6 output channels.


\begin{table}[]
    \centering
        \caption{The Average True Match Rate (TMR) @ 10\% False Match Rate (FMR) of our method compared to current state-of-the-art methods under a common evaluation protocol. Note that the training is performed on morphs created from synthetically generated face images. }
    \label{tab:tmr}
    \resizebox{\columnwidth}{!}{
    \begin{tabular}{|c|c|c|c|c|c|}
        \hline
         Dataset & \textbf{diffDeMorph (ours)}&\textbf{SDeMorph} \cite{ref18} & \textbf{IPD} \cite{ref66} & \textbf{Face Demorphing} \cite{ref51}  \\
          \cline{1-5}
         AMSL&  99.49\%&45.32\%& 72.36\%&22.13\% \\
         OpenCV& 100.00\%&20.54\%& 68.74\%&21.72\% \\

         FaceMorpher& 100.00\%&57.01\% &67.69\% &20.05\%\\
         WebMorph& 99.84\%&55.12\%&65.52\% &18.18\% \\
         MorDiff& 99.81\%&53.76\%& 63.22\%&18.33\% \\
         StyleGAN&  99.32\%&30.54\% & 7.32\%& 7.02\%\\
         
         \hline
         
    \end{tabular}}

\end{table}

\vspace{-0.3cm}
\section{Datasets}
\vspace{-0.3cm}
A significant bottleneck in reference-free face demorphing is the limited availability of large-scale face morph datasets due to privacy concerns. Existing morphing datasets, primarily designed for Morph Attack Detection (MAD), typically contain around 1,000 morphs, which is insufficient for training data-hungry generative models.
To address this limitation, we train our method on morphs generated from synthetically generated faces, which also allows us to address the privacy concerns associated with using real biometric data. We test the method on real morph datasets to assess its performance in a more practical, real-world use case.
\begin{table}[]
    \centering
    \caption{Comparison of our baseline (same train and test dataset) with the current state-of-the-art demorphing methods in terms of TMR at 10\% FMR, when both the train and test morphs originate from the same dataset, i.e., using a consistent morph technique. We report the scores as presented in \cite{ref51}.  }
    
    \resizebox{0.9\columnwidth}{!}{
    \begin{tabular}{|c|c|c|c|c|}
    \hline
         Dataset& \textbf{diffDeMorph (Ours)} &\textbf{SDeMorph} \cite{ref18} & \textbf{IPD} \cite{ref66} & \textbf{Face Demorphing} \cite{ref51} \\
         \hline
         AMSL       &73.67\%   &12.56\% & 25.69\% & 70.55\%\\
         OpenCV     & 88.73\%  &-     & -     &-\\
         FaceMorpher& 96.59\%  & 13.18\%& 37.82\%&-\\
         WebMorph   & 82.04\%  & 12.80\%& 25.61\%&-\\
         MorDiff    & 100.00\%  & 11.67\%& 38.12\%&-\\
         StyleGAN   & 59.50\%  & 0.00\% & 16.22\%&-\\
         \hline
    \end{tabular}}
    \label{tab:baseline-comp}
\end{table}
\textbf{Training Dataset}: 
During training, we randomly sample two face images from SMDD \cite{ref20} bonafide set and create a morph using the widely adopted \cite{sarkar2020vulnerabilityanalysisfacemorphing,9093905} OpenCV/dlib morphing algorithm \cite{ref97}, based on Dlib's landmark detector implementation \cite{ref98}. We generate 15,000 morphs for training using the face images from the SMDD dataset's training set. All images are processed with MTCNN \cite{ref71} to detect faces, and only the detected face regions are cropped. The images are then normalized and resized to a resolution of $256 \times 256$. Images where faces cannot be detected are discarded. Importantly, no additional spatial transformations are applied, ensuring that the facial features (such as the lips and nose) of both the morphs and the ground-truth constituent images remain properly aligned during training.
\textbf{Testing Dataset}: We evaluate our  method on  three widely used morph datasets: AMSL \cite{ref64}, FRLL-Morphs \cite{ref65}, and MorDiff \cite{ref9}. The FRLL-Morphs dataset consists of morphs generated using four different techniques: OpenCV \cite{ref97}, StyleGAN \cite{ref69}, WebMorph \cite{ref70}, and FaceMorph \cite{ref68}. In all three datasets, the source (non-morph) images are drawn from the FRLL dataset, which contains 102 identities, each represented by two frontal images—one smiling and one neutral—resulting in a total of 204 face images. The number of morphs in each dataset is as follows: AMSL: 2,175 morphs; FaceMorpher: 1,222 morphs; StyleGAN: 1,222 morphs; OpenCV: 1,221 morphs; WebMorph: 1,221 morphs; MorDiff: 1,000 morphs. The selected test datasets include both traditional landmark-based techniques and more recent generative methods.

\begin{figure}
    \centering
    \includegraphics[width=0.7\linewidth]{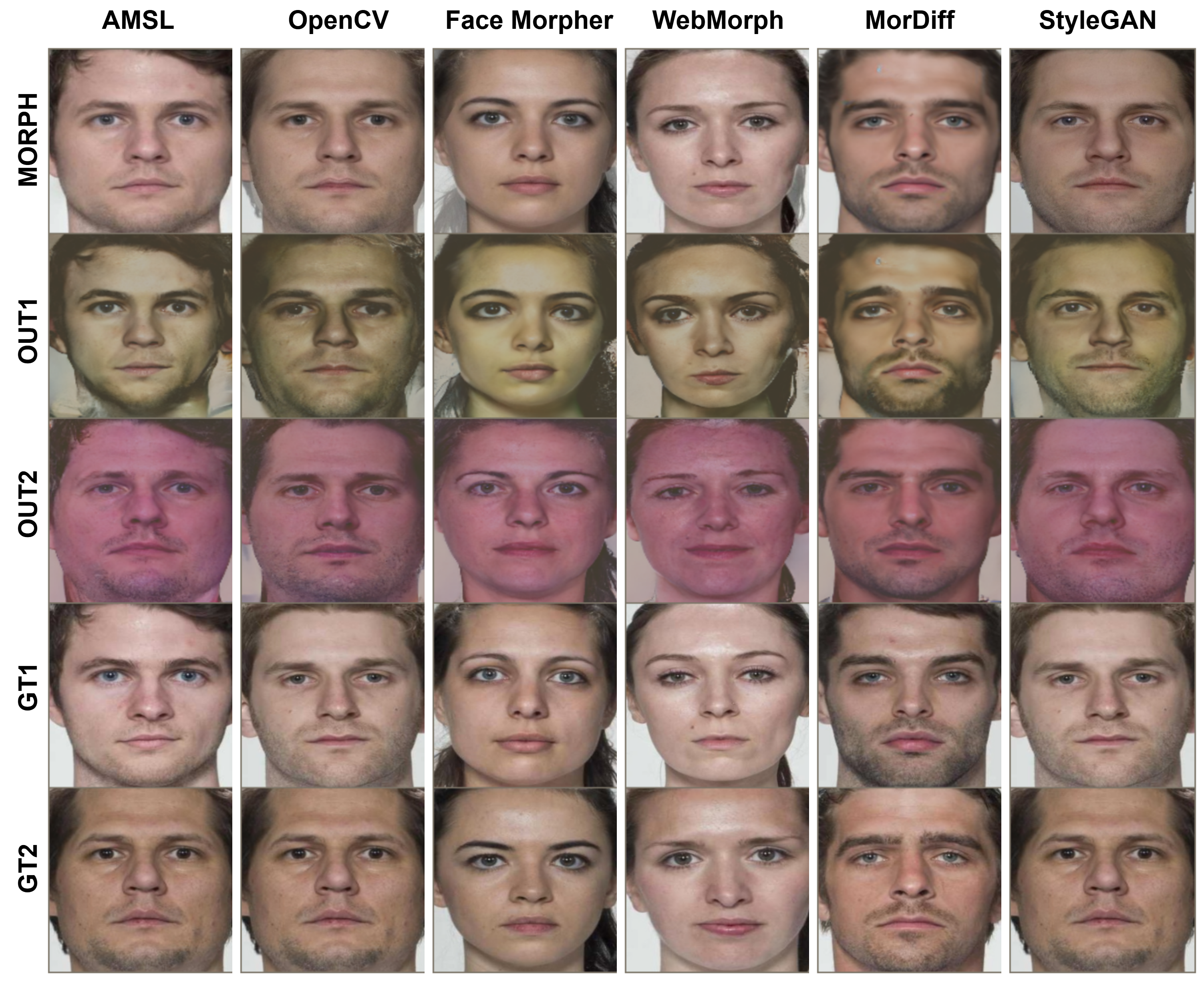}
    \caption{Samples generated by our method under an assumption-free training protocol. Our method generates i) distinct looking faces with higher visual fidelity compared to previous methods and, ii) the generated face images match the ground truth more faithfully.   }
    \label{fig:samples}
\end{figure}

\begin{table}[]
    \centering
    \caption{Comparing diffDeMorph against current state-of-the-art methods. 
    Our method outperforms existing methods by a considerable margin in terms of 
    Restoration Accuracy (RA) in scenario 3.}

    \resizebox{0.45\textwidth}{!}{%
    \begin{tabular}{|l|c|c|c|c|c|c|c|c|}
        \hline
        \textbf{Dataset} 
        & \multicolumn{2}{|c|}{\textbf{diffDeMorph (Ours)}} 
        & \multicolumn{2}{|c|}{\textbf{SDeMorph} \cite{ref18}} 
        & \multicolumn{2}{|c|}{\textbf{IPD} \cite{ref66}} 
        & \multicolumn{2}{|c|}{\textbf{Face Demorphing} \cite{ref51}} \\ 
        \cline{2-9}
         & AdaFace & ArcFace 
        & AdaFace & ArcFace 
         & AdaFace & ArcFace 
        & AdaFace & ArcFace \\ 
        \hline
        AMSL        &  51.69\% & 99.49\%  & 0.00\% & 12.56\% & 0.18\%  & 25.69\% & 0.17\% & 0.45\% \\ 
        OpenCV      & 58.88\% & 100.00\%  & 0.00\% & 15.62\%  & 1.89\% & 40.54\%  & 0.23\% & 0.53\% \\ 
        FaceMorpher & 59.79\% & 100.00\%  & 0.00\% & 13.18\%  & 1.43\% & 37.82\%  & 0.17\% & 0.51\% \\ 
        WebMorph    & 40.26\% & 99.82\%   & 0.00\% & 12.80\% & 0.31\%  & 25.61\%  & 0.20\% & 0.50\% \\ 
        Mordiff     & 66.13\% & 100.00\%  & 0.00\% & 11.67\% & 3.88\%  & 38.12\%  & 0.29\% & 0.62\% \\ 
        StyleGAN    & 67.00\% & 95.97\%   & 0.00\% & 0.00\%  & 0.0\%   & 16.22\%  & 0.00\% & 0.43\% \\ 

        \hline
    \end{tabular}%
    }
    \label{tab:metrics}
\end{table}
\vspace{-0.3cm}
\section{Experiments and Results}
\vspace{-0.3cm}
\label{sec:results}
\textbf{Evaluation Protocol and Metrics: } Given a face morph $x$, created using constituent  images $i_1,i_2$, the denoiser outputs $o_1$ and $o_2$. We pair the outputs with the ground truth face images, $i_1$ and $i_2$,  by finding the similarity score for the two possible output-to-ground truth pairs. We use a face comparator $\mathcal{B}$ to assess facial similarity. If  $\mathcal{B}(\text{$o_1$}, \text{$i_1$}) + \mathcal{B}(\text{$o_2$}, \text{$i_2$})$ is greater than $\mathcal{B}(\text{$o_1$}, \text{$i_2$}) + \mathcal{B}(\text{$o_2$}, \text{$i_1$})$, we pair $o_1$ with $i_1$ and $o_2$ with $i_2$; otherwise, we pair $o_1$ with $i_2$ and $o_2$ with $i_1$. The genuine score is computed by finding the facial similarity between the output and the ground truth image it is paired with.
The impostor score is computed by identifying the closest matching face in the FRLL face image database, excluding the ground truth images. To evaluate our method, we use the True Match Rate (TMR) at 10\% False Match Rate (FMR) and Restoration Accuracy (RA). The similarity threshold while calculating TMR depends on the FMR, whereas in the case of RA, it is fixed beforehand (typically 0.4).  As for the biometric matcher $\mathcal{B}$, we employ two publicly available  face matchers, namely, AdaFace \cite{ref22} and ArcFace \cite{ref77}.
\begin{table}[htbp]
    \centering
    \caption{Performance of diffDeMorph when the training and testing is done on same dataset.
    }
    \resizebox{0.7\columnwidth}{!}{%
    \begin{tabular}{|l|c|c|c|c|}
        \hline
        \textbf{Dataset} 
        & \textbf{PSNR} 
        & \textbf{SSIM} 
        & \multicolumn{2}{|c|}{\textbf{RA / TMR}} \\ 
        \cline{4-5}
        & & & \textbf{Adaface} & \textbf{Arcface} \\ 
        \hline
        AMSL        & 9.04  & 0.33 & 0.00/2.65 & 71.55/73.67 \\ 
        OpenCV      & 9.39  & 0.35 & 1.41/4.22 & 88.73/88.73 \\ 
        FaceMorpher & 10.20 & 0.36 & 6.82/24.43 & 96.52/96.59 \\ 
        WebMorph    & 9.38  & 0.35 & 0.00/2.81 & 82.04/82.04 \\ 
        MorDiff     & 9.73  & 0.35 & 1.67/13.37 & 97.50/100 \\ 
        StyleGAN    & 8.66  & 0.32 & 0.00/0.00 & 63.38/59.5 \\ 
        
        \hline
    \end{tabular}%
    }
    \label{tab:baseline-tmr}
\end{table}
The protocol in scenario 3 can be further divided  into three categories : train and test morphs i) are generated using the same morph technique; ii) have the same face style (e.g., passport style, similar background etc.); and iii) are completely independent. Note that the third category reflects a real-world use and is the most challenging. Our main contribution in this paper is a demorphing technique which is free of any assumptions on train and test morphs, i.e., the third category.\vspace{-0.3cm}
\subsection{Demorphing with consistent morph technique}
\vspace{-0.2cm}
Firstly, we consider the `consistent morph technique' baseline. In other words, we train and test our method on the same dataset. The identities in each dataset are split using an identity-disjoint protocol with a 60-40 train-test ratio. Morphs are included in the train (respectively, test)  set only if both identities belong to the train (test) set. Morphs with identities spanning both sets (i.e., one in each) are removed. To ensure consistency, only neutral facial expressions are considered. Our method achieves an average TMR@10\%FMR of 73.67\% on AMSL dataset compared to 70.55\% in \cite{ref51}, 12.56\% in \cite{ref18} and 25.69\% in \cite{ref66}.  Our baseline results consistently outperform current methods when tested in scenario 3. We present the results in Table \ref{tab:baseline-comp} and Table \ref{tab:baseline-tmr}.\vspace{-0.2cm}
\subsection{Demorphing with consistent face style}
\vspace{-0.2cm}
We also evaluate our method under the assumption that the train and test morphs are created from \textit{passport-style} face images. We train our method on the combined train sets from the six datasets and test on the combined test sets accordingly. Under this assumption, our method achieves RA of 60.88\% and TMR@10\%FMR of 61.25\%, which outperforms existing methods by a considerable margin.\vspace{-0.2cm}
\subsection{Generalized Demorphing }
\vspace{-0.2cm}
Finally, we relax all the assumptions regarding the distribution of train and test morphs. Note that this is the most challenging scenario and reflects a real-world case. We train on morphs generated from synthetic face images and test on real morph datasets.  We present the results in Table \ref{tab:tmr} and Table \ref{tab:metrics}\footnote{To compute the Restoration Accuracy (RA), we use a fixed matching threshold of 0.4, following prior works. However, this fixed threshold leads to reduced RA on AdaFace due to the presence of artifacts, as illustrated in Figure \ref{fig:comparison}.} and visualize them in Figure \ref{fig:samples}. On the AMSL dataset, our method outperforms its closest competitor in terms of restoration accuracy (RA) by 73.80\%. These figures for the remaining datasets are 59.46\%, 62.18\%, 74.21\%, 61.88\% and 79.75\%.  In terms of TMR@10\%FMR, our method outperforms the closest existing methods by 27.13\%, 31.26\%, 32.31\%, 34.32\%, 36.59\% and 68.78\% across the six datasets evaluated.
\begin{figure}
    \centering
    \includegraphics[width=0.7\linewidth]{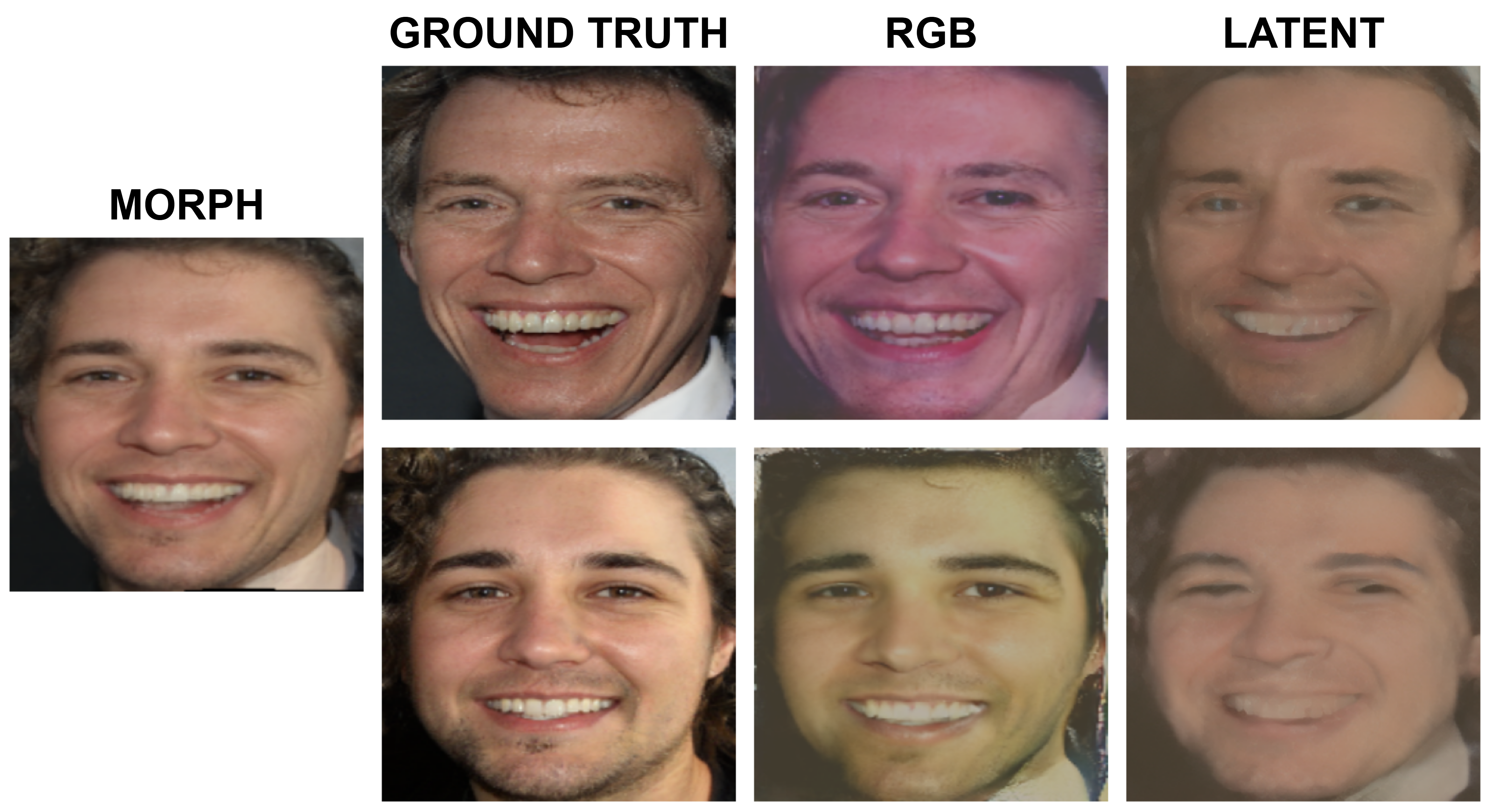}
    \caption{Effects of conditioning in RGB space and latent space. Compression in latent space removes seemingly unimportant details like hair style, background etc. which are crucial for demorphing. }
    \label{fig:enter-label}
\end{figure}

\noindent\textbf{Ablation Study: Guidance in RGB vs latent domain.} We performed identical experiments in the latent domain to evaluate the effectiveness of conditioning in the RGB space. On average, conditioning in the RGB domain results in a TMR of 99.21\% and RA of 99.97\% across all six datasets, outperforming the latent domain, which achieves 87.72\% and 97.44\% with respect to TMR and RA, respectively. This difference is due to the compression in the latent domain, which eliminates details (hair, background, etc.) in the morph images that, while not important for tasks such as morph-detection, are crucial for demorphing. 
\vspace{-0.3cm}
\section{Conclusion}
\vspace{-0.3cm}
We propose a novel generative face demorphing method that makes limited assumptions about the train and test morphs. Our method uses DDPMs conditioned on the morph image in the RGB domain to reconstruct the constituent images. We also conducted various experiments  based on the morph technique and image style. Our method outperforms  existing methods by a considerable margin under a common more realistic training protocol.
\vspace{-0.5cm}

{\small
\balance
\bibliographystyle{ieee_fullname}
\bibliography{refs}
}
\end{document}